\pdfoutput=1

\documentclass[11pt]{article}

\usepackage[preprint]{acl}

\usepackage{times}
\usepackage{latexsym}
\usepackage{xcolor}
\usepackage{colortbl}

\usepackage[T1]{fontenc}

\usepackage[utf8]{inputenc}

\usepackage{microtype}

\usepackage{inconsolata}

\usepackage{graphicx}
\usepackage{amsmath}
\usepackage{times}
\usepackage{framed}
\usepackage{ragged2e}
\usepackage{bm}
\usepackage{soul}
\usepackage{latexsym}
\usepackage{subfigure}
\usepackage{amsthm}
\usepackage{graphicx}
\usepackage{float}
\usepackage{longtable}
\usepackage{booktabs} 
\usepackage{multirow}
\usepackage{multicol}
\usepackage{amssymb}
\usepackage{dashbox}%
\usepackage{multirow}
\usepackage{textcomp}
\usepackage{tcolorbox}
\usepackage{bbding}
\usepackage{bbm}
\usepackage[ruled, vlined, linesnumbered]{algorithm2e}
\usepackage{mathtools}
\usepackage{adjustbox}
\usepackage[ruled,vlined]{algorithm2e}
\usepackage{color, soul}
\usepackage{pifont}
\usepackage{array}
\newcolumntype{P}[1]{>{\centering\arraybackslash}p{#1}}
\newcolumntype{M}[1]{>{\centering\arraybackslash}m{#1}}
\usepackage{cleveref}
\crefname{section}{§}{§§}
\Crefname{section}{§}{§§}
\crefname{figure}{Figure}{Figure}
\Crefname{figure}{Figure}{Figure}
\crefname{table}{Table}{Table}
\Crefname{table}{Table}{Table}
\usepackage{tabularx}
\newcommand\ourmethod{\textsc{VerIF}\xspace}
\newcommand\ourdata{\textsc{VerInstruct}\xspace}

\title{\ourmethod: Verification Engineering for Reinforcement Learning \\ in Instruction Following}

\author{Hao Peng, Yunjia Qi, Xiaozhi Wang, Bin Xu, Lei Hou, Juanzi Li\\
Department of Computer Science and Technology, Tsinghua University \\
\texttt{\{peng-h24\}@mails.tsinghua.edu.cn}}

\begin{document}
\maketitle
\begin{abstract}

Reinforcement learning with verifiable rewards (RLVR) has become a key technique for enhancing large language models (LLMs), with verification engineering playing a central role. 
However, best practices for RL in instruction following remain underexplored. In this work, we explore the verification challenge in RL for instruction following and propose \ourmethod, a verification method that combines rule-based code verification with LLM-based verification from a large reasoning model (e.g., QwQ-32B). To support this approach, we construct a high-quality instruction-following dataset, \ourdata, containing approximately 22,000 instances with associated verification signals. We apply RL training with \ourmethod to two models, achieving significant improvements across several representative instruction-following benchmarks. The trained models reach state-of-the-art performance among models of comparable size and generalize well to unseen constraints. We further observe that their general capabilities remain unaffected, suggesting that RL with \ourmethod can be integrated into existing RL recipes to enhance overall model performance. We have released our datasets, codes, and models to facilitate future research\footnote{\url{https://github.com/THU-KEG/VerIF}}.

\end{abstract}

\section{Introduction}

Reinforcement learning with verifiable rewards (RLVR) has emerged as a key technique for enhancing large language models (LLMs), leading to various advanced LLMs, such as DeepSeek R1~\citep{guo2025deepseek}. The core component of RLVR is verification engineering. Recently, numerous works have explored reliable verification across diverse domains, such as math~\citep{lambert2024t, guo2025deepseek, deepscaler2025}, code~\citep{wang2024enhancing, deepcoder2025}, logic~\citep{xie2025logic}, medicine~\citep{chen2024huatuogpt,wang2025baichuan}, and finance~\citep{qian2025fino1,liu2025fin}.

\begin{figure}[htb]
    \centering
    \includegraphics[width=0.95\linewidth]{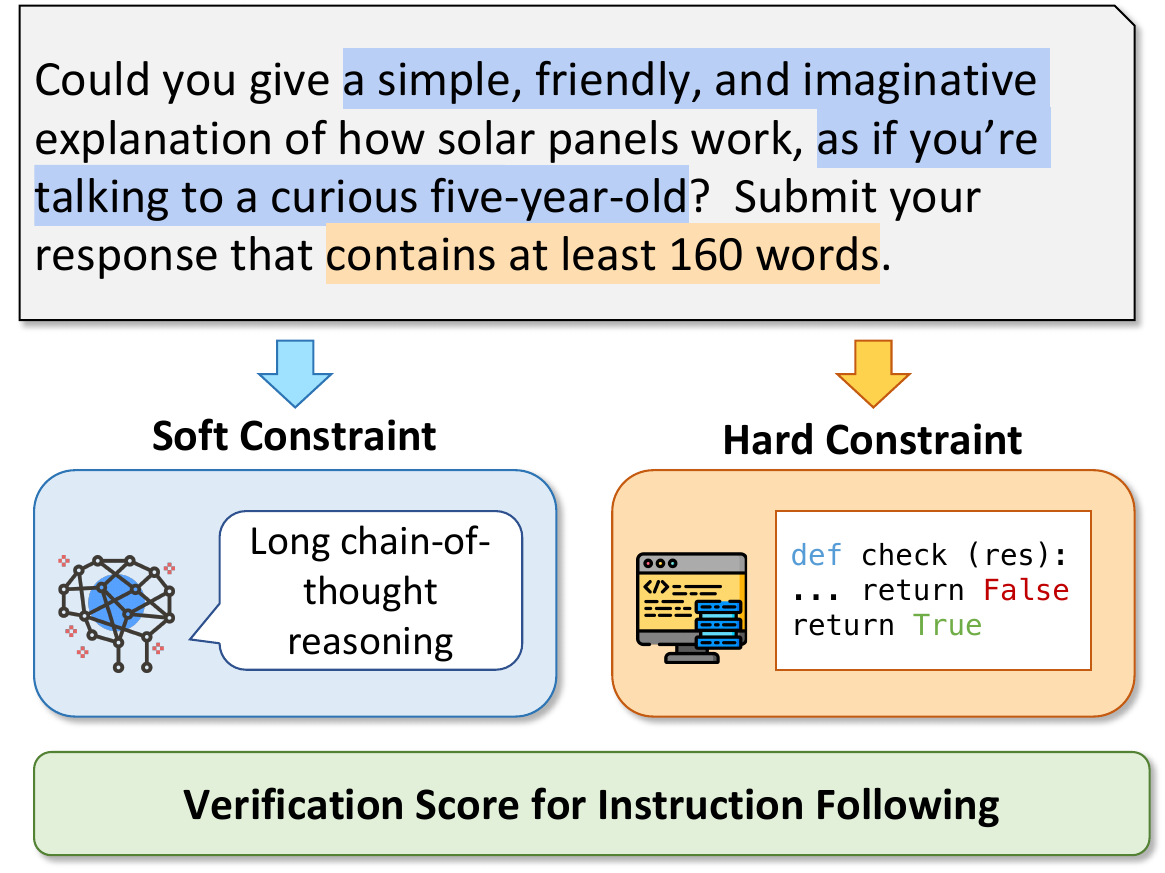}
    \caption{A simplified illustration of \ourmethod. The instruction constraints are categorized as soft or hard and verified using different methods in \ourmethod.}
    \label{fig:fig1}
\end{figure}

In this work, we explore verification engineering for reinforcement learning in instruction following. 
Specifically, this work focuses on the following of constraints in the instruction~\citep{zhou2023instruction}, such as response length, as shown in Figure~\ref{fig:fig1}.
The constraints are usually divided into two types: \textit{hard} constraints, which can be verified using simple rules, e.g., length, and \textit{soft} constraints, which require semantic judgment, e.g., style. Assessing whether a response satisfies these constraints provides a natural basis for verification in RLVR. However, reinforcement learning for instruction following remains underexplored. The only notable work, TULU 3~\citep{lambert2024t}, applies RLVR to enhance instruction following. However, the improvement is limited, and it focuses solely on hard constraints, neglecting soft constraints. Therefore, the best practice of verification engineering for RL in instruction following remains under-explored.

Given the above issues, we explore the best practice of RLVR in instruction following and propose \ourmethod, a verification method for instruction following that combines rule-based code verification with verification from a large reasoning model. As shown in Figure~\ref{fig:fig1}, hard constraints are verified through code, and soft constraints are handled by a large reasoning model, which enables effective verification through long chain-of-thought reasoning~\citep{liu2025inference}. \ourmethod requires no manual annotations or reference answers, offering an efficient solution for automatic verification. To support this approach, we construct a high-quality dataset, \ourdata, containing approximately $22,000$ instances with verification. The data construction involves two main steps: (1) instruction construction with multiple constraints, where we apply constraint back-translation~\citep{qi2024constraint} to augment existing instructions with additional constraints; (2) verification generation. For hard constraints such as length, we use Qwen2.5-72B-Instruct~\citep{yang2024qwen2} to generate verification code. For soft constraints, they are verified online during RL training using large reasoning models.

We apply reinforcement learning with \ourmethod on two SFT-trained models using \ourdata, including TULU 3 SFT~\citep{lambert2024t} and DeepSeek-R1-Distill-Qwen-7B~\citep{guo2025deepseek}. Specifically, \ourmethod computes the final reward as the average of hard constraint scores (0 or 1) from code validation and soft constraint scores (0 or 1) determined by the QwQ-32B~\citep{QwQ32B}. We train the models using the GRPO algorithm~\citep{Shao2024DeepSeekMathPT}. We evaluate the trained models on several widely-used instruction-following benchmarks, including IFEval~\citep{zhou2023instruction}, Multi-IF~\citep{He2024MultiIFBL}, SysBench~\citep{qin2024sysbench}, FollowBench~\citep{jiang2024followbench}, and CFBench~\citep{zhang2024cfbench}. Experimental results show that the RLVR-trained models using \ourmethod achieve significant improvements. Notably, the model trained based on TULU 3 SFT achieves state-of-the-art performance among models of similar parameter scale and outperforms TULU 3~\citep{lambert2024t}, which is trained with extensive DPO data and rule-based RLVR. 
The results demonstrate the effectiveness of our verification method \ourmethod.

We conduct further analytical experiments. We first evaluate the generalization of the trained models on general instruction following tasks, including AlpacaEval 2.0~\citep{dubois2024length} and MT-Bench~\citep{zheng2023judging},  and mathematical reasoning tasks, including GSM8K~\citep{cobbe2021training} and Omni-MATH~\citep{gaoomni25}, natural language understanding datasets: MMLU-Pro~\citep{wang2024mmlu} and DROP~\citep{dua2019drop}, and a natural language inference benchmark BBH~\citep{suzgun2023challenging}. We observe that RL with \ourmethod preserves general and mathematical capabilities, indicating its potential as an additional RL stage to enhance instruction following without affecting other skills. We analyze the performance gains of trained models across different constraint types and find that RL with \ourmethod exhibits good generalization to unseen constraints. We also conduct ablation studies on the verification method, using only code validation or only LLM verification, both of which lead to notable performance drops. Finally, we develop a smaller and efficient 7B LLM as the \textit{soft} constraint verifier. Specifically, we extract approximately 130k complex instructions from WildChat~\citep{zhaowildchat24} and Infinity Instruct~\citep{InfinityInstruct2024}, collect responses from 6 different LLMs, and use QwQ to generate constraint verification. We then train DeepSeek-R1-Distill-Qwen-7B on this dataset as a generative verifier for \textit{soft} constraints, achieving RL performance comparable to the model trained using QwQ-32B as the verifier.



\section{Pilot Experiments}
This section explores the potential of RL for instruction following (\cref{sec:potential}) and preliminarily explores different verification methods (\cref{sec:verification_engineering}) using the reward benchmark IFBench~\citep{peng2025agentic}.

\subsection{Potential for RL Training}
\label{sec:potential}
\begin{figure}[t]
    \centering
    \includegraphics[width=0.93\linewidth]{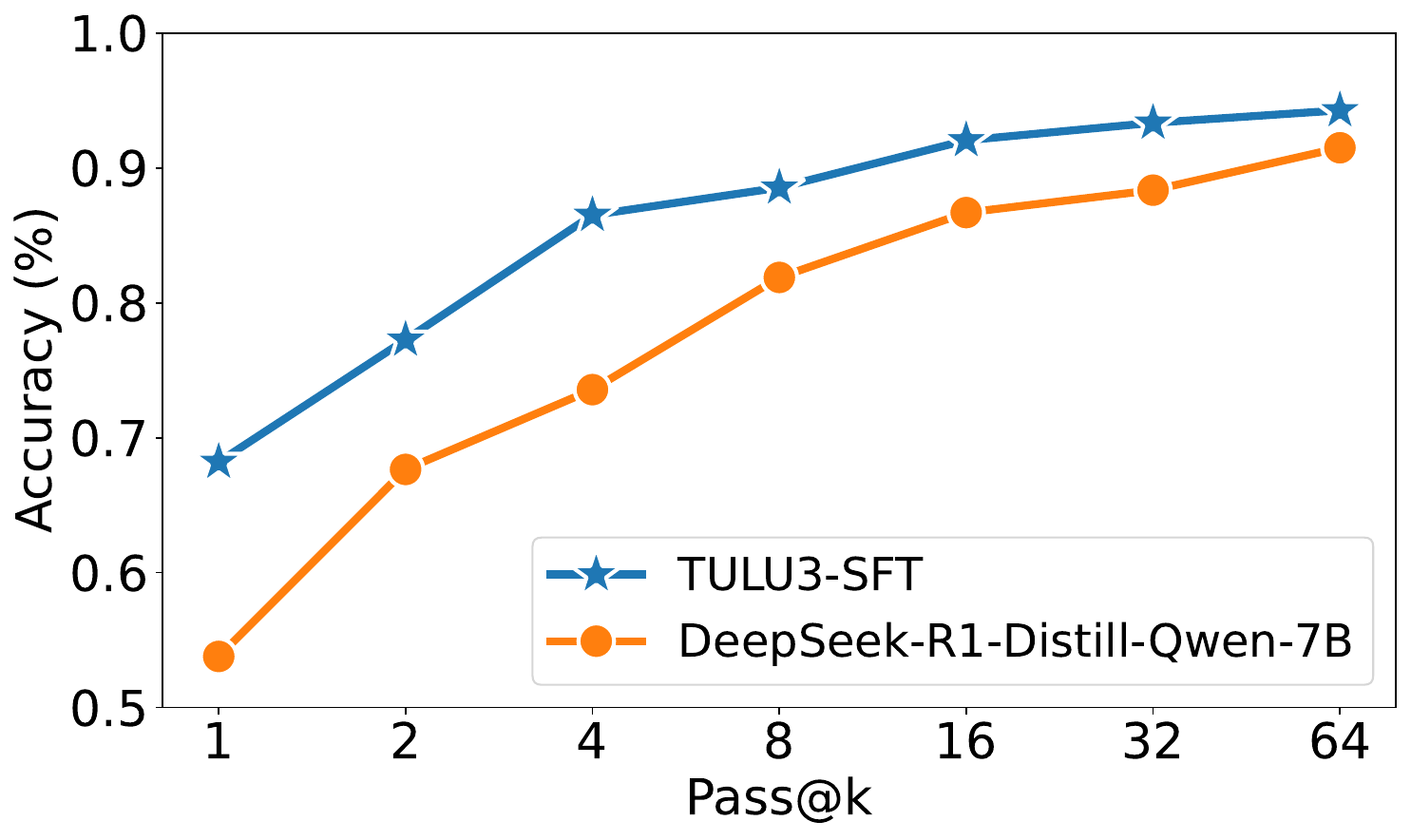}
    \caption{Pass@$k$ results (\%) of two SFT-trained LLMs on IFEval. We report the prompt-level strict score.}
    \label{fig:pass_n}
\end{figure}

We first explore the potential of RL in instruction following, as most previous works have adopted supervised fine-tuning (SFT; \citealp{ouyang2022training}) or direct preference optimization (DPO; ~\citealp{rafailov2023direct}), with limited use of RL. This raises a key question: \textit{Does RL hold untapped potential for instruction following}? To explore this question, we evaluate the pass@$k$ performance of several LLMs on the instruction following benchmark IFEval~\citep{zhou2023instruction}. The motivation is that RL enhances performance by increasing the likelihood of sampling correct responses, and a high pass@$k$ at large $k$ suggests untapped potential that RL can exploit~\citep{yue2025does}. 
The experimental results of TULU 3 SFT and DeepSeek-R1-Distill-Qwen-7B are shown in Figure~\ref{fig:pass_n}. We can observe that the results are much higher with larger $k$, with pass@$64$ showing over a $20\%$ increase compared to pass@$1$. This suggests that LLMs can sample correct answers on IFEval at higher $k$, with the potential that can be exploited during RL training.

\subsection{Verification Engineering}
\label{sec:verification_engineering}

\begin{table}
    \small
    \centering
    \begin{tabular}{lccc}
    \toprule
    Method & Hard & Soft & Overall \\ \midrule
    Code-only & $60.6$  & $13.2$ & $48.6$ \\
    LLM-only\textsubscript{\textsc{QwQ}} & $31.5$ & $48.1$ & $37.4$ \\
    LLM-only\textsubscript{\textsc{Qwen}} & $19.7$ & $45.3$ & $28.6$ \\
    Code+LLM\textsubscript{\textsc{QwQ}}& $61.3$ & $48.1$ & $58.1$ \\
    \bottomrule
    \end{tabular}
    \caption{Accuracy (\%) of three verification methods on IFBench. ``Hard'' or ``Soft'' indicates that the rejected response only violates certain hard or soft constraints.}
    \label{tab:pilot}
\end{table}

\begin{figure}[t]
    \centering
    \includegraphics[width=0.95\linewidth]{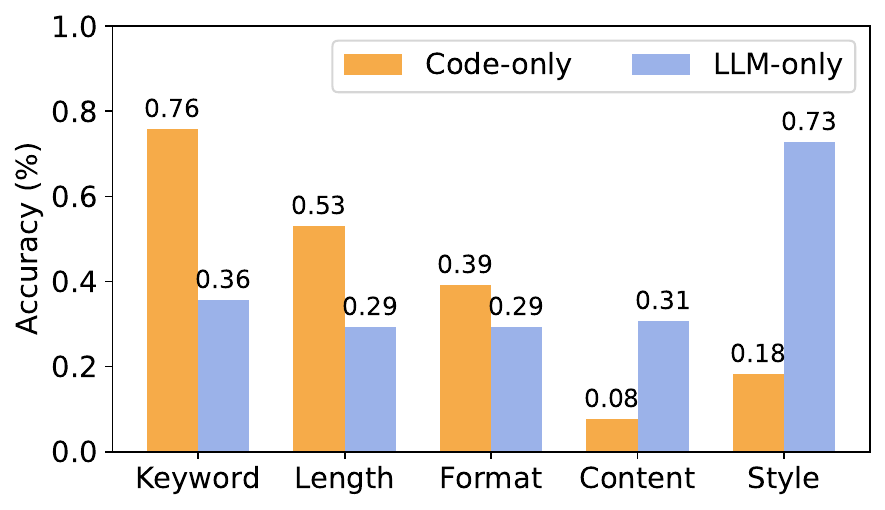}
    \caption{Accuracy (\%) of code-only or LLM-only verification in verifying compliance with different types of constraints. LLM-only adopts QwQ-32B.}
    \label{fig:constraint_type}
\end{figure}

We preliminarily conduct verification engineering using reward model benchmarks. Specifically, we evaluate different verification methods on IFBench~\cite{peng2025agentic}, a benchmark designed for instruction-following rewards that consists of an instruction and two responses, where the task is to select the response that better follows the instruction. IFBench includes $3$ common hard constraints: length, format, and keyword, and $2$ common soft constraints: style and content.
We explore three verification methods: (1) code-only verification, similar to RewardAgent proposed by \citet{peng2025agentic}, which uses automatically generated code for each constraint verification; (2) LLM-only verification, which directly uses the LLM as the judge; (3) code+LLM verification, which applies code verification for hard constraints and LLM for soft constraints. We explore using QwQ-32B~\citep{QwQ32B} and Qwen2.5-72B-Instruct~\citep{yang2024qwen2} as the LLMs. The results are shown in Table~\ref{tab:pilot}. We can observe that code+LLM verification performs much better and reasoning LLMs (QwQ) also perform better than non-reasoning LLMs (Qwen).

We further investigate the accuracies of code and LLM verification on different types of constraints, and report their respective accuracy for soft and hard constraints in Table~\ref{tab:pilot}, which further confirms that code verification is more effective for hard constraints and LLM verification performs better on soft constraints, supporting the rationale for the code+LLM verification approach. The detailed results across different constraint types are shown in Figure~\ref{fig:constraint_type}, and we can observe that LLMs perform particularly poorly on keyword and length constraints, which may be due to inherent limitations in numerical counting~\citep{fu2024large,ballcan24}. Since keyword and length constraints can be efficiently verified with code, we conclude that in instruction-following verification, hard constraints should be checked with code, and soft constraints can be reliably verified by advanced LLMs.

\section{Method}
This section introduces the formalization of \ourmethod (\cref{sec:verif}), the construction process of \ourdata (\cref{sec:verinstruct}), and the RL training method (\cref{sec:rl_training}).

\begin{figure*}[t]
    \centering
    \includegraphics[width=0.95\linewidth]{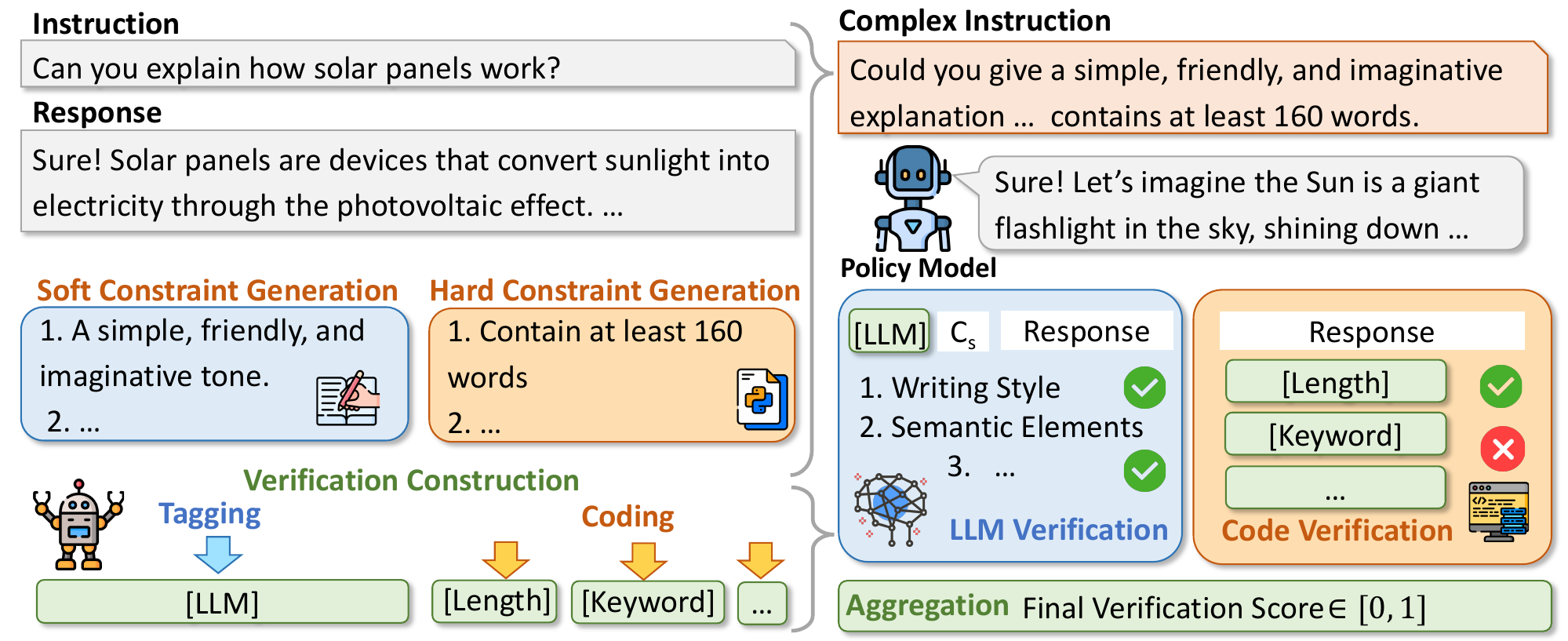} 
    \caption{\textbf{Left:} The construction process for \ourdata, including complex instruction generation and verification construction. \textbf{Right:} Our verification method, \ourmethod, providing verification for instruction following.
    }
    \label{fig:workflow}
\end{figure*}

\subsection{Verification Method}
\label{sec:verif}
Suppose we are given an instruction 
$x$, which includes the task decription and a set of constraints $C=\{c_1, c_2, ..., c_n\}$. We follow the task definition of instruction-following by \citet{zhou2023instruction}: given $x$, generating a response $y$ that satisfies all constraints in $C$. In this work, our primary goal is to accurately verify whether $y$ meets all constraints and to apply this reliable verification in reinforcement learning training. Specifically, the constraint set $C$ consists of two types: hard constraints $C_h$, which can be verified by simple rules or code (e.g., length), and soft constraints $C_s$, which require semantic understanding (e.g., style). As explored in \cref{sec:verification_engineering}, we propose a hybrid verification approach, \ourmethod, that uses code verification for $C_h$ and LLM verification for $C_s$. Formally, this is defined as:
$$
\label{eq:eq_1}
\text{\ourmethod}(x, y) = F(\text{Code}(y, C_h), \text{LLM}(y, C_s))
$$
$\text{Code}(y, C_h) \in \{0, 1\}$ denotes whether $y$ satisfies all hard constraints in $C_h$, and $\text{LLM}(y, C_s) \in \{0, 1\}$ indicates whether $y$ satisfies all soft constraints in $C_s$. $F$ denotes the aggregation method used to combine the code verification score and the LLM verification score, including averaging or multiplication. In this work, we consider only three types of hard constraints, including length, format, and keyword. All other constraints are taken as soft and verified using LLMs. As explored in \cref{sec:verification_engineering}, we use large reasoning models for LLM-based verification, which is a form of scaling up verification and has been demonstrated effective in practice~\citep{liu2025inference, seed2025seed15thinkingadvancingsuperbreasoning}.

\subsection{Data Construction Method}
\label{sec:verinstruct}
We construct a high-quality instruction-following dataset for reinforcement learning, where each instance is paired with a corresponding verification. Prior works on enhancing instruction-following of LLMs~\citep{sun2024conifer, dong2024self, qi2024constraint} have primarily focused on generating complex instructions and corresponding high-quality responses for supervised fine-tuning (SFT). 
In this work, we focus on generating complex instructions with associated verification, eliminating the efforts to generate and filter high-quality responses. As shown in Figure~\ref{fig:workflow}, the construction process consists of two main parts: (1) \textbf{Complex instruction generation.}
We adopt the constraint back-translation approach~\citep{qi2024constraint} to generate complex instructions, which produces few unrealistic cases. Specifically, we randomly sample $25,000$ data instances from four high-quality datasets, including Alpaca GPT4~\citep{alpaca-gpt4}, Orca Chat~\citep{Orca-Chat}, Evol Instruct~\citep{xu2023wizardlm}, and OpenAssitant~\citep{kopf2024openassistant}. We use Llama3.1-70B-Instruct~\citep{grattafiori2024llama} to generate constraints implicitly satisfied by each response, such as language style. Since LLMs often struggle with understanding length constraints~\citep{sun2024conifer}, we instead automatically synthesize them based on response length using Python scripts.
We combine the generated constraints with the original instruction to form the final complex instruction.
 (2) \textbf{Verification construction.}
We then automatically generate a verification method for each constraint. For hard constraints, including length, format, and keyword presence, we use Qwen2.5-72B-Instruct to generate verification Python code. Given the simplicity of these generated Python code scripts, we manually check them and find nearly no errors. For soft constraints, we do not generate code but instead tag them with ``LLM'', which indicates that verification during RL training should be online produced by an LLM.
We finally filter out instructions with fewer than $2$ constraints, resulting in \ourdata, which contains $22,000$ instructions, each including an average of $6.2$ constraints and corresponding verification methods. The details of \ourdata are placed in Appendix~\ref{sec:app_data}.

\begin{table*}[t]
    \centering
    \small
    \resizebox{\linewidth}{!}{
    \begin{tabular}{lcccccccccc}
    \toprule
    \multirow{2}{*}{Model} & \multicolumn{4}{c}{IFEval} & \multicolumn{3}{c}{Multi-IF} & SysBench & FollowBench & CFBench \\
    \cmidrule{2-11} 
    & Pr. (S) & Pr. (L) & Ins. (S) & Ins. (L) & Turn 1 & Turn 2 & Turn 3 & ISR & SSR & ISR \\
    \midrule
    GPT-4o&$79.9$&$84.8$&$85.6$&$89.6$&$82.3$&$71.7$&$59.3$&$80.2$&$75.3$&$80.0$\\
    QwQ-32B&$82.8$&$86.1$&$88.0$&$90.4$&$64.2$&$56.6$&$48.4$&$67.8$&$73.5$&$80.0$\\
    Qwen2.5-7B-Instruct&$71.5$&$74.1$&$79.4$&$81.3$&$75.3$&$57.9$&$47.0$&$-$&$65.9$&$74.0$\\
    LLaMA3.1-8B-Instruct&$72.6$&$77.3$&$80.8$&$84.2$&$71.3$&$62.8$&$54.6$&$-$&$65.9$&$71.0$\\
    TULU 3&$79.7$&$82.8$&$85.1$&$87.5$&$82.1$&$63.2$&$51.2$&$48.9$ &$70.3$&$72.0$\\
    \midrule
    Crab-7B-DPO&$47.3$&$57.1$&$59.7$&$67.9$&$47.2$&$36.5$&$28.9$&$-$&$56.3$&$62.0$\\
    Conifer-7B-DPO&$48.1$&$52.3$&$59.1$&$63.3$&$50.7$&$37.6$&$26.6$&$-$&$56.9$& $62.0$\\
    UltraIF-8B-DPO$\dagger$&$71.3$&$75.4$&$79.4$&$83.1$&$69.6$&$58.3$&$46.9$&$-$&$62.6$&$-$\\
    \midrule
    R1-Distill-Qwen-7B &$59.9$&$65.1$&$70.4$&$74.2$&$55.8$&$43.6$&$32.7$&$16.9$&$53.9$&$66.0$\\
    \rowcolor{cyan!20}
    \quad +\ourmethod &$75.6$&$79.5$&$82.7$&$85.5$&$66.0$&$53.8$&$41.9$&$26.5$&$61.0$&$68.0$\\
    TULU 3 SFT &$68.4$&$71.7$&$76.3$&$79.5$&$67.3$&$50.9$&$40.3$&$33.2$&$62.0$&$63.0$\\
    \rowcolor{cyan!20}
    \quad +\ourmethod &$84.5$&$87.1$&$89.3$&$91.4$&$79.4$&$65.2$&$54.0$&$54.7$&$68.6$&$72.0$\\
    
    \bottomrule
    \end{tabular}
    }
    \caption{Experimental results (\%) on several representative instruction-following benchmarks. ``Pr.'' and ``Ins.'' denote prompt-level and instruction-level metrics respectively. ``S'' and ``L'' mean strict and loose respectively. $\dagger$ denotes the results are sourced from the original paper~\citep{an2025ultraif}. All the other results are reproduced by us in this paper. For reasoning LLMs, we remove the thinking tokens and evaluate using only the final response.}
    \label{tab:main_exp}
\end{table*}

\subsection{RL Training}
\label{sec:rl_training}

We conduct reinforcement learning using \ourmethod on \ourdata. Specifically, we adopt the GRPO algorithm~\citep{Shao2024DeepSeekMathPT} and perform $16$ rollouts per prompt for value estimation. For each response, the reward is provided online by \ourmethod. To reduce the overhead of LLM-based verification, we input all soft constraints $C_s$ to the LLM at once to assess whether the response satisfies all of them in a single pass. We conduct RL training using the VeRL framework\footnote{\url{https://github.com/volcengine/verl}} and integrate a parallel reward computation mechanism to accelerate RL training.

\section{Experiments}
This section introduces experimental setup (\cref{sec:setup}), main results (\cref{sec:main_results}), analytical experiments (\Cref{sec:general,sec:constraint_type,sec:ablation}), and developing a smaller verifier (\cref{sec:small_verifier}).


\subsection{Experimental Setup}
\label{sec:setup}
\paragraph{Reported Models}
We conduct RL training based on two SFT-trained models: TULU 3 SFT~\citep{lambert2024t} and DeepSeek-R1-Distill-Qwen-7B~\citep{guo2025deepseek}. For the specific implementation of \ourmethod, we use QwQ-32B as the LLM verifier and set $F$ in Equation~\ref{eq:eq_1} as average. For comparison, we evaluate TULU 3~\citep{lambert2024t}, which is trained directly based on TULU 3 SFT with extensive DPO and RLVR training. We also evaluate various industrial models, including GPT-4o~\citep{hurst2024gpt}, QwQ-32B~\citep{QwQ32B}, Qwen2.5-7B-Instruct~\citep{yang2024qwen2}, LLaMA3.1-8B-Instruct~\citep{grattafiori2024llama}, and open-source models specifically optimized for instruction following, including Conifer~\citep{sun2024conifer}, Crab~\citep{qi2024constraint}, and UltraIF~\citep{an2025ultraif}. More details are placed in appendix~\ref{sec:app_detail}.

\paragraph{Evaluation benchmarks}
We evaluate the models on several representative instruction-following benchmarks, including IFEval~\citep{zhou2023instruction}, the most commonly used dataset; Multi-IF~\citep{He2024MultiIFBL}, which includes multi-turn and multilingual instruction following; SysBench~\citep{qin2024sysbench}, which evaluates instruction following to system prompts; FollowBench~\citep{jiang2024followbench} and CFBench~\citep{zhang2024cfbench}, which cover a comprehensive range of constraint types.



\begin{table*}[t]
    \centering
    \small
    \begin{tabular}{lccccccc}
    \toprule
    Model & AlpacaEval 2.0 & MT-Bench & GSM8K & Omni-MATH & MMLU-Pro & BBH & DROP \\
    \midrule
    Qwen2.5-7B-Instruct&$37.5$&$7.8$&$91.4$&$13.6$&$56.5$&$71.8$&$77.2$\\
    Llama3.1-8B-Instruct&$29.4$&$6.0$&$83.6$&$10.8$&$48.1$&$63.0$&$74.4$\\
    TULU 3 &$39.9$&$7.5$&$88.4$&$14.2$&$35.9$&$68.5$&$69.4$\\
    \midrule
    R1-Distill-Qwen-7B&$16.6$&$5.7$&$87.0$&$35.0$&$54.3$&$21.5$&$74.0$\\
    \rowcolor{cyan!20}
    \quad +\ourmethod &$15.5$&$5.9$&$90.0$&$33.6$&$54.8$&$32.2$&$75.6$\\
    TULU 3 SFT &\phantom{0}$7.9$&$6.3$&$78.8$&$11.4$&$36.4$&$67.4$&$58.3$\\
    \rowcolor{cyan!20}
    \quad +\ourmethod &$22.0$ &$7.0$&$83.4$&$12.4$&$36.0$&$67.9$&$59.5$\\
    
    \bottomrule
    \end{tabular}
    \caption{Experimental results (\%) on various general natural language benchmarks.}
    \label{tab:general}
\end{table*}

\subsection{Main Results}
\label{sec:main_results}

All experimental results are presented in Table~\ref{tab:main_exp}. We have the following observations:
(1) Reinforcement learning with \ourmethod demonstrates strong performance. Compared to their corresponding backbones (R1-Distill-Qwen-7B and TULU 3 SFT), the trained models using RL perform much better. Notably, the model trained based on TULU 3 SFT even outperforms the original TULU 3~\citep{lambert2024t}, which is trained based on TULU 3 SFT using approximately $271$k DPO pairs and specialized RLVR data. Among models with similar parameter scales, the model trained based on TULU 3 SFT achieves state-of-the-art performance and surpasses several open-source models developed by industry using larger datasets and more resources. This demonstrates the potential of RL training for instruction following and the effectiveness of \ourmethod in providing reliable rewards.
(2) RL with \ourmethod generalizes effectively to unseen instruction-following tasks. Although the training dataset \ourdata contains only English and single-turn instruction-following data, the trained model shows substantial improvements on multilingual, multi-turn (Multi-IF) instruction following, and following system prompts (SysBench). This suggests that the patterns of instruction following may be inherently generalizable and that RL further enhances this generalization~\citep{chu2025sft}.
(3) RL with \ourmethod benefits both reasoning and non-reasoning models. As reinforcement learning has demonstrated its effectiveness in enhancing reasoning abilities on challenging tasks~\citep{guo2025deepseek,deepcoder2025}, such as math and code, we suggest integrating instruction-following training into RL pipelines. Our further analysis (\cref{sec:general}) shows that general capabilities, such as mathematical reasoning and language understanding, do not degrade after RL with \ourmethod and may even slightly improve, indicating that RL with \ourmethod can be integrated into broader model development for enhancing the model's instruction following capabilities.
(4) Models developed by the academic community, such as Conifer, Crab, and UltraIF, show relatively lower performance, which is reasonable given their focus on exploring effective SFT data synthesis and limited training resources. Given that there is abundant open-source SFT data, such as Infinity Instruct~\citep{InfinityInstruct2024} with approximately 7 million instances, we encourage the research community to devote more attention to constructing RL data instead, as RL data remains scarce and RL has been demonstrated to be effective for instruction following. In conclusion, RL with \ourmethod effectively enhances instruction-following capabilities, and we encourage more efforts on developing effective RL methods or data for instruction-following.

\subsection{Analysis on General Capabilities}
\label{sec:general}
We further investigate the general capabilities of the trained models to assess the broader impact of RL with \ourmethod. 
Specifically, we conduct an evaluation on various representative general benchmarks, including general instruction-following datasets that focus on task completion and are evaluated using LLM-as-a-judge: AlpacaEval 2.0~\citep{dubois2024length} and MT-Bench~\citep{zheng2023judging}, mathematical reasoning benchmarks: GSM8K~\citep{cobbe2021training} and Omni-Math~\citep{gaoomni25}, natural language understanding datasets: MMLU-Pro~\citep{wang2024mmlu} and DROP~\citep{dua2019drop}, and a natural language inference benchmark BBH~\citep{suzgun2023challenging}.
The results are shown in Table~\ref{tab:general}. We can observe that RL training does not degrade general performance and even improves the performance in some cases, such as MT-Bench, GSM8K, and BBH. 
We attribute this to a key difference between RL and SFT: while SFT learns and memorizes patterns from data and is prone to catastrophic forgetting~\citep{chu2025sft}, RL typically maximizes optimal patterns it has learned~\citep{yue2025doesreinforcementlearningreally}, thereby reducing the risk of knowledge forgetting.
These results suggest a promising finding that instruction-following reinforcement learning can be integrated into existing RL pipelines to enhance adherence to instructions without compromising the model’s general capabilities.


\subsection{Analysis on Constraint Types}
\label{sec:constraint_type}

\begin{figure}[t]
    \centering
    \includegraphics[width=1.0\linewidth]{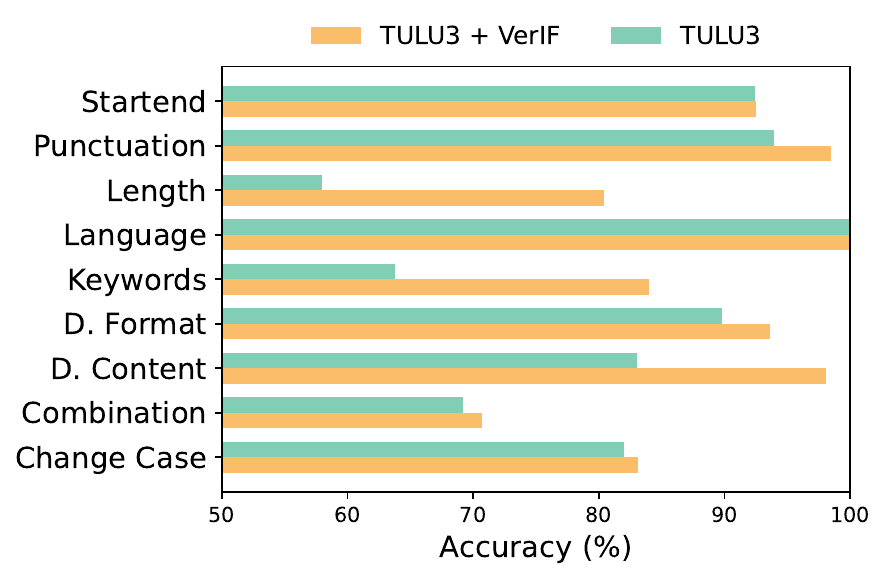}
    \caption{Prompt-level strict scores (\%) across different types of constraints on IFEval. ``D.'' denotes Detectable.}
    \label{fig:ifeval}
\end{figure}

\ourdata includes only five constraint types: length, keyword, format, content, and style. We further investigate the improvements across different constraint types in IFEval to analyze the generalization of constraint adherence. Results are shown in Figure~\ref{fig:ifeval}. We observe clear improvements across most types except for ``Startend'' and ``Language'' (which already achieves $100\%$ accuracy). This indicates that RL training can generalize instruction-following ability to unseen constraint types. For constraint types covered in \ourdata, such as length, keyword, and content, the improvements are more pronounced, which demonstrates the precision of the verification provided by \ourmethod. This also suggests that incorporating datasets with richer constraint types can further improve performance. We encourage the community to explore more diverse data for RL for instruction following.

\subsection{Ablation Studies}
\label{sec:ablation}

\begin{figure}[t]
    \centering
    \includegraphics[width=1.0\linewidth]{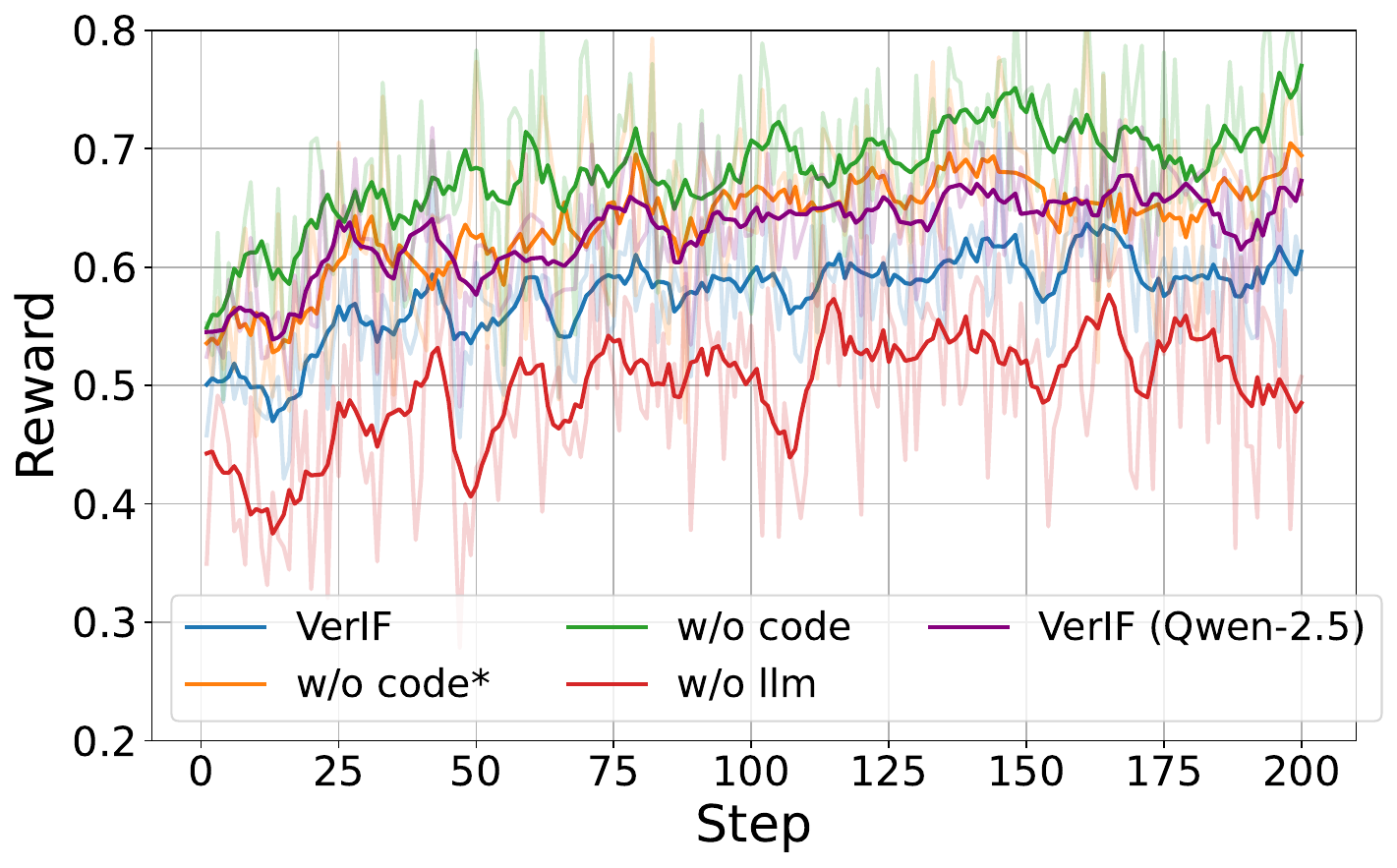}
    \caption{Reward curves during RL training with different verification methods. We visualize the first $200$ steps and smooth the data for better visualization.}
    \label{fig:ablation}
\end{figure}


\begin{table}[]
    \centering
    \small
    \begin{tabular}{lcccc}
    \toprule
    Model & IFEval & Multi-IF & CFBench \\
    \midrule
    \ourmethod&$84.5$&$54.0$&$72.0$\\
    \quad w/o code* &$81.7$&$51.2$&$70.0$\\
    \quad w/o code &$76.2$&$52.0$&$73.0$\\
    \quad w/o LLM &$74.7$&$46.0$&$59.0$\\
    \midrule
    \ourmethod (Qwen-2.5) &$76.9$&$48.5$&$71.0$\\
    \bottomrule
    \end{tabular}
    \caption{Ablation results (\%) for different verification methods.  ``Qwen-2.5'' uses Qwen2.5-72B-Instruct as the LLM instead of QwQ-32B. We report the prompt-level strict score for IFEval, the Turn 3 score for Multi-IF, and the ISR score for CFBench.}
    \label{tab:ablation}
\end{table}

\begin{table*}[t]
    \centering
    \small
    \resizebox{\linewidth}{!}{
    \begin{tabular}{lcccccccccc}
    \toprule
    \multirow{2}{*}{Model} & \multicolumn{4}{c}{IFEval} & \multicolumn{3}{c}{Multi-IF} & SysBench & FollowBench & CFBench \\
    \cmidrule{2-11} 
    & Pr. (S) & Pr. (L) & Ins. (S) & Ins. (L) & Turn 1 & Turn 2 & Turn 3 & ISR & SSR & ISR \\
    \midrule
    \ourmethod (QwQ-32B)&$84.5$&$87.1$&$89.3$&$91.4$&$79.4$&$65.2$&$54.0$&$54.7$&$68.6$&$72.0$\\
    \midrule
    \ourmethod (Qwen-7B)&$77.1$&$80.4$&$84.3$&$86.6$&$78.5$&$60.8$&$49.0$&$42.7$&$62.0$&$72.0$\\
    \rowcolor{cyan!20}
    \ourmethod (IF-Verifier-7B)&$80.0$&$84.5$&$86.0$&$89.4$&$80.1$&$63.7$&$52.7$&$49.5$&$68.8$&$70.0$\\
    
    \bottomrule
    \end{tabular}
    }
    \caption{Experimental results (\%) of models trained using different LLM verifiers. Qwen-7B is short for DeepSeek-R1-Distilled-Qwen-7B. The base model used for RL training is TULU 3 SFT.}
    \label{tab:small_verifier}
\end{table*}


We conduct ablation studies on the verification method. Specifically, we perform three ablations: (1) ``w/o code*'', which uses only the LLM to verify all constraints; (2) ``w/o code'', which uses only the LLM for soft constraints; (3) ``w/o LLM'', which verifies only hard constraints using Python code scripts. We conduct RL training using different verification methods based on TULU 3 SFT. 
The reward curves during training are shown in Figure~\ref{fig:ablation}. We observe that using only code verification yields lower rewards and limited growth, likely due to the difficulty of following hard constraints. In contrast, using only LLM verification results in higher and more pronounced reward growth, possibly because the LLM verifier is easier to fit or hack~\citep{li2024generation}. The results are shown in Table~\ref{tab:ablation}. We can observe that removing any verification component degrades model performance compared to \ourmethod. Notably, ``w/o LLM'', which uses only Python scripts for hard constraint verification, performs significantly poorly. This may be due to that approximately $77.7\%$ constraints in training data are soft. This suggests that using code verification for hard constraints only, as adopted in training TULU 3~\citep{lambert2024t}, is sub-optimal for RL in instruction following. We also adopt Qwen2.5-72B-Instruct as the LLM verifier in \ourmethod and find it significantly underperforms QwQ-32B. The potential reason may be that in our implementation of \ourmethod, the LLM is required to verify whether a response satisfies all soft constraints in a single pass, which requires step-by-step reasoning and poses significant challenges. The results demonstrate the potential of scaling up verification~\citep{liu2025inference}. In conclusion, we suggest that the best practice for verification in RL for instruction following is \ourmethod with reasoning models as the LLM verifier.
We further explore a smaller reasoning LLM as the verifier in \cref{sec:small_verifier}.

\subsection{Training a Smaller Verifier}
\label{sec:small_verifier}

\begin{figure}[t]
    \centering
    \includegraphics[width=1.0\linewidth]{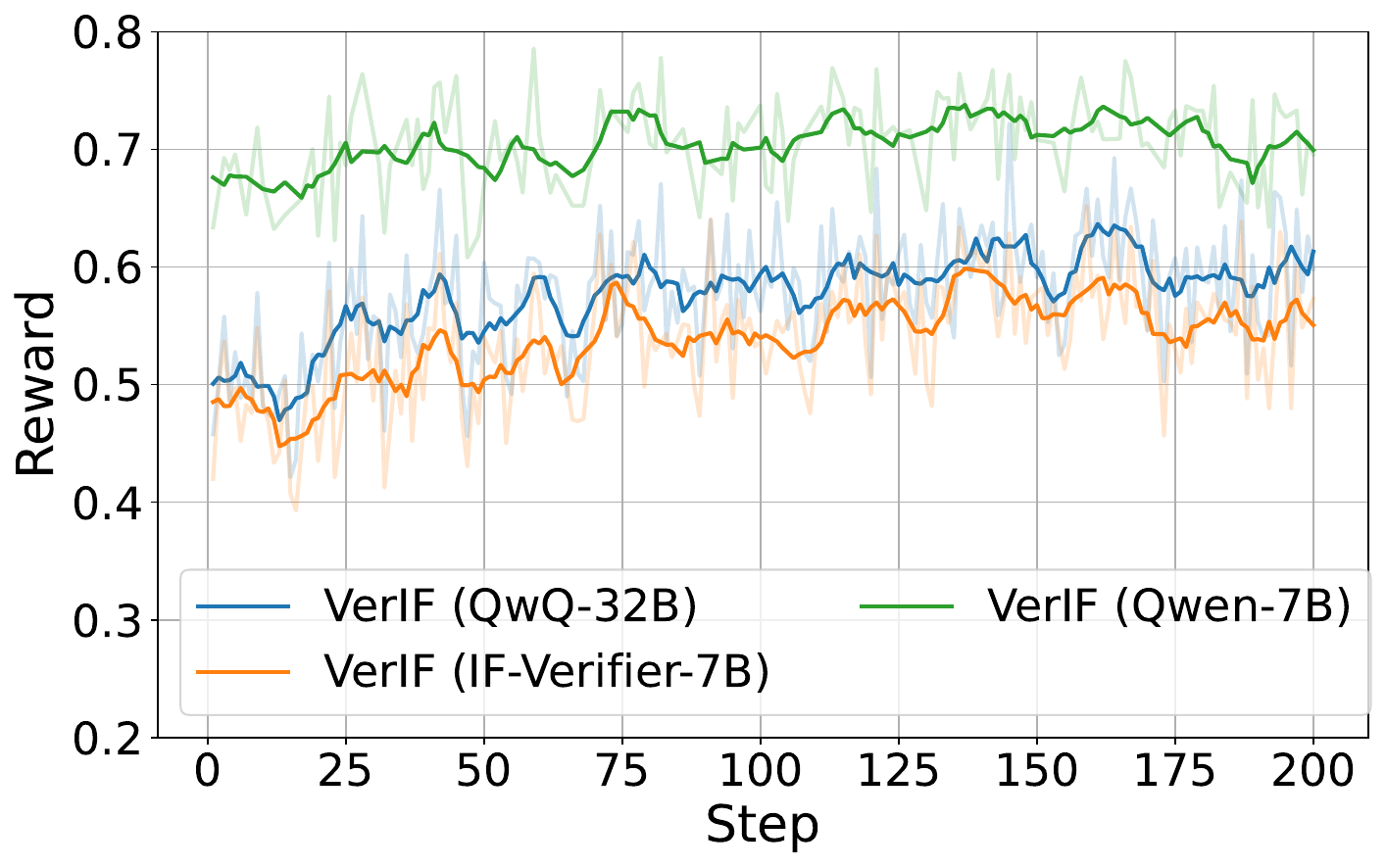}
    \caption{Reward curves during RL training using different LLM verifiers in \ourmethod. Qwen-7B is short for DeepSeek-R1-Distilled-Qwen-7B.}
    \label{fig:small_reward}
\end{figure}

Although we have demonstrated \ourmethod with a large reasoning model, such as QwQ-32B, is effective for RL in instruction following, the long outputs of QwQ-32B lead to high latency during online reward computation. For example, when training TULU 3 SFT, we adopt 8 H800 GPUs for deploying QwQ-32B and set batch size to 32, rollouts to 16, and the average time to obtain the reward for a batch reaches about $180$ seconds, accounting for roughly $80\%$ of the time per training step. To address this, we explore using smaller reasoning models as LLM verifiers while maintaining comparable performance. A straightforward approach is to distill a verifier from QwQ-32B. Therefore, we distill $130$k SFT data instances from QwQ, where each instance consists of an instruction, a response, and a critic indicating whether a response satisfies the given constraints in the instruction. The data collection process is detailed in Appendix~\ref{sec:app_verifier}.

We fine-tune DeepSeek-R1-Distill-Qwen-7B on the collected dataset, resulting in \textbf{IF-Verifier-7B}. We then conduct RL training on TULU 3 SFT using the new LLM verifiers in \ourmethod. Figure~\ref{fig:small_reward} shows the reward curves during training. We can observe that DeepSeek-R1-Distill-Qwen-7B yields higher initial rewards, but its reward growth is limited. IF-Verifier-7B exhibits a 
similar reward trajectory as QwQ-32B. The results of the trained models are shown in Table~\ref{tab:small_verifier}. We observe that using IF-Verifier-7B as the LLM verifier in \ourmethod significantly outperforms DeepSeek-R1-Distill-Qwen-7B and achieves competitive performance to QwQ-32B. Moreover, IF-Verifier-7B reduces computational cost a lot. Deploying IF-Verifier-7B 
requires only one single H800 GPU, with an average reward computation time of $120$ seconds per batch, which can be further reduced with multi-GPUs. This makes \ourmethod a practical method for effective RL training under limited resources. This work preliminarily explores more efficient LLM verifiers and encourages further efforts~\citep{liu2025inference}.


%


\section{Related Work}
Instruction following requires models to generate responses that satisfy complex user instructions. Recent work has primarily focused on following constraints in instructions, such as length and keyword~\citep{zhou2023instruction}. Existing efforts to enhance instruction-following capabilities primarily focus on methods for (1) collecting SFT data, including directly distilling from larger LLMs~\citep{sun2024conifer, he2024complex,dong2024self,ren2025step}, back-translation~\citep{qi2024constraint, pham2024suri}, and training dedicated instruction composers~\citep{an2025ultraif}, and (2) collecting preference pairs~\citep{cheng2024spar, pham2024suri, dong2024self,zhang2024iopo}. Notably, two works are similar to ours: AutoIF~\citep{dong2024self}, which constructs both instructions and corresponding verification code, but but does not explore RL training or soft constraints verification; and TULU 3~\citep{lambert2024t}, which adopts RLVR for instruction following but the improvement is limited and also does not consider soft constraints. In summary, the best practice for RL in instruction following remains underexplored.


As RL has proven to be an effective post-training technique, many prior studies have explored its applications across various domains, primarily focusing on verification engineering, such as math~\citep{lambert2024t, guo2025deepseek, deepscaler2025,seed2025seed15thinkingadvancingsuperbreasoning}, code~\citep{wang2024enhancing, deepcoder2025}, logic~\citep{xie2025logic}, tool using~\citep{feng2025retool, jin2025search, qian2025toolrl, li2025torl,zheng2025deepresearcher}, machine translation~\citep{wang2024drt,he2025r1}, medicine~\citep{chen2024huatuogpt,wang2025baichuan}, and finance~\citep{qian2025fino1,liu2025fin}. In this work, we explore the best practice of RL for instruction following and propose \ourmethod, an effective verification method for RL training.

\section{Conclusion}
In this work, we propose \ourmethod, an effective verification method for RL in instruction following. We also construct \ourdata, a dataset for instruction following where each instruction is paired with corresponding verification signals. We perform RL training with \ourmethod on \ourdata, leading to significant improvements. The trained models achieve SoTA performance on several representative instruction-following benchmarks at a similar model scale, without hurting general capabilities. This work demonstrates the promising potential of RL in instruction following, and we encourage further exploration of novel RL methods and data.
\section*{Limitations}


We discuss the limitations of our work here, including two main aspects: (1) The training dataset \ourdata includes only English data, which may limit the broader usage of the dataset. We observe that RL on \ourdata still generalizes well to multiple languages, and we encourage the community to collect more diverse data covering more languages. (2) \ourmethod relies on an LLM as the verifier, which inherits common issues of LLM-as-a-judge, such as potential biases~\citep{ye2024justice} and vulnerability to adversarial attacks~\citep{shi2024optimization}. We believe developing more robust and efficient LLM judges~\citep{liu2025inference} is a promising direction and leave it for future work.

\section*{Ethical Considerations}

We discuss potential ethical concerns as follows:
(1) Intellectual property. 
Alpaca-GPT4 and Infinity-Instruct are licensed under CC BY-NC 4.0\footnote{\url{https://creativecommons.org/licenses/by-nc/4.0/}}, OpenAssistant is licensed under Apache License 2.0\footnote{\url{https://www.apache.org/licenses/LICENSE-2.0}}. WildChat is licensed under ODC-By license\footnote{\url{https://opendatacommons.org/licenses/by/1-0/}}. Evol-Instruct and Orca-Chat do not specify explicit licenses. We strictly adhered to all claimed licenses. Our dataset will be released under the Apache License 2.0. We believe the original open-source datasets are properly anonymized, and we do not introduce any additional sensitive information. 
(2) Potential risk control. In this paper, we propose \ourmethod, a verification method for RL to improve instruction-following capabilities of LLMs. As \ourmethod includes an LLM verifier, it inherits the known risks of LLMs, such as potential bias~\citep{gallegos2024bias, ye2024justice}. We do not introduce any additional risks. Users should not exploit \ourmethod for reward hacking~\citep{skalse2022defining} and are responsible for verifying the compliance of the models trained using it. (3) AI assistance. We use ChatGPT and Claude to refine some sentences.

\bibliography{custom}
\newpage
\clearpage
\appendix
\section*{Appendices}

\section{\ourdata}
\label{sec:app_data}


\subsection{Dataset Construction Details}
Following~\citet{qi2024constraint}, we collect original instructions and responses from several publicly available instruction-tuning datasets, including Alpaca GPT-4~\citep{alpaca-gpt4}, Orca Chat~\citep{Orca-Chat}, Evol Instruct~\citep{xu2023wizardlm}, and OpenAssistant~\citep{kopf2024openassistant}. We then apply a back-translation-based method to extract both soft and hard constraints from the instruction-response pairs. 
Table~\ref{tab:constraint_generation_prompt} presents the prompt to generate soft constraints used in \ourdata construction.
The hard constraints, including Length, Keyword, and selected aspects of Format, are automatically generated through Python-based processing.
These constraints are subsequently combined to form the final constraint-enhanced prompt.

\subsection{Dataset Statistics}

\begin{figure}[t]
    \centering
    \includegraphics[width=1.0\linewidth]{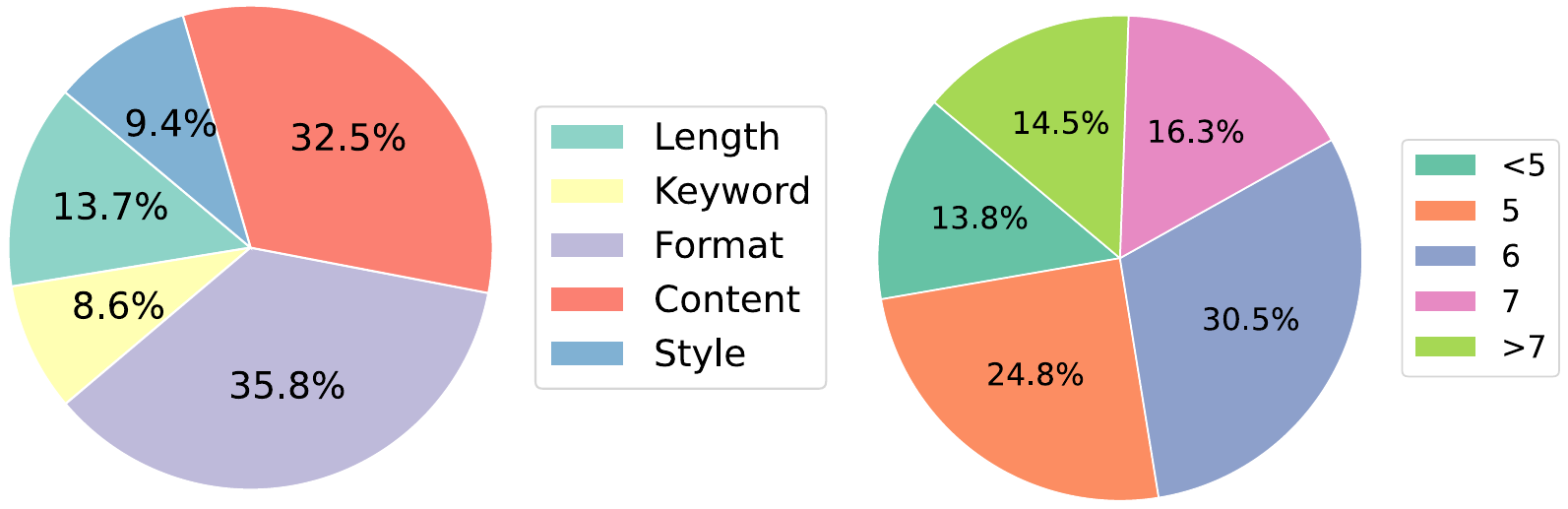}
    \caption{
    \textbf{left:} Proportional distribution of constraint types in the dataset.
\textbf{right:} Distribution of the number of constraints per instruction.
    }
    \label{fig:constraint_info}
\end{figure}

Figure~\ref{fig:constraint_info} shows the distribution of $22,000$ instances in the \ourdata dataset.
Following IFBench~\citep{peng2025agentic}, we categorize constraints into five types: length, keyword, format, content, and style.
The left chart presents the proportional distribution of constraint types.
Since certain format constraints, such as those requiring hierarchical output structures, are not easily verifiable via Python, we define format, content, and style as soft constraints, which together account for $77.7\%$ of the total.
Length and keyword are defined as hard constraints, making up the remaining $22.3\%$.
The right chart categorizes the data by the number of constraints after merging.

\section{Experimental Details}
\label{sec:app_detail}

We train our model using the open-source VeRL framework\footnote{\url{https://github.com/volcengine/verl}} with the GRPO algorithm~\citep{Shao2024DeepSeekMathPT}, setting the KL loss coefficient to $1 \times 10^{-3}$. The batch size is set to $32$, the number of rollouts to $16$, the maximum generation length of rollout to $4,096$, and the learning rate to $1 \times 10^{-6}$. We save checkpoints every 20 steps during training. Following TULU 3~\citep{lambert2024t}, we use IFEval~\citep{zhou2023instruction} as the validation set to select the best checkpoint. We train the models for one epoch on \ourdata with early stopping if performance on IFEval does not improve for more than 5 checkpoints. The best checkpoints are typically found within the first $200$ steps. For evaluation, we set the sampling temperature to 0 to ensure reproducibility. For all evaluations using LLM-as-a-judge, we adopt gpt-4o-2024-11-20 as the judge. Since the Conifer model~\citep{sun2024conifer} is not publicly open-sourced, we instead train a model using its SFT and DPO data, and the reported results of Conifer are evaluated based on our reproduced model. For the evaluation of reasoning LLMs, we remove thinking tokens and evaluate only the final responses. For evaluation of general capabilities, we report the length-controlled win rate for AlpacaEval 2.0~\citep{dubois2024length}. Both training and evaluation are conducted on Nvidia H800 GPUs, with the entire training process taking approximately $1,900$ GPU hours in total.



\section{Training a Small Verifier}
\label{sec:app_verifier}

We provide a detailed description of the training data construction process and training details. Following the construction of \ourdata, we first generate an additional $20,000$ data instances. To ensure diversity, we additionally mined complex instructions from WildChat~\citep{zhaowildchat24} and Infinity Instruct~\citep{InfinityInstruct2024}. Specifically, we use Qwen2.5-72B-Instruct to extract constraints from each instruction and classify them as hard or soft. For hard constraints, we adopt Qwen2.5-72B-Instruct to generate corresponding verification Python code scripts. The full prompt is presented in Table~\ref{tab:classify_constraint}. For each instruction, we randomly sample a response from $6$ different models, including Llama3.1-8B-Instruct~\citep{grattafiori2024llama}, Llama-3.3-70B-Instruct~\citep{grattafiori2024llama}, Qwen2.5-7B-Instruct~\citep{yang2024qwen2}, Qwen2.5-72B-Instruct~\citep{yang2024qwen2}, QwQ-32B~\citep{QwQ32B}, DeepSeek-R1-Distilled-Qwen-32B~\citep{guo2025deepseek}. We then adopt QwQ-32B to generate a step-by-step verification indicating whether the output satisfies the instruction for each instruction-response pair. As a result, we collect about $130$k instruction–response pairs with corresponding step-by-step verification. For SFT training, we use the open-source alignment-handbook framework~\citep{Tunstall_The_Alignment_Handbook}. Based on DeepSeek-R1-Distill-Qwen-7B, we train the model on the collected dataset for $2$ epochs, with $2 \times 10^{-5}$ learning rate, $64$ batch size, $8,192$ max sequence length, resulting in the verifier IF-Verifier-7B.


%

\begin{table*}[t]
    \centering
    \small
    \begin{adjustbox}{max width=\linewidth}
    \begin{tabular}{p{\linewidth}}
    \toprule
    \textbf{Prompt: Generating Constraints from Instruction and Output} \\[0.5em]

    As a linguist with expertise in contextual language nuances, please add constraints to enrich the \#Given Instruction\# based on the \#Given Output\#. The goal is to enhance the specificity and detail of the instruction to ensure that the response is more aligned with the output text. \\[0.5em]

    To supplement the instruction using the output, please consider adding \textbf{specific and detailed} constraints across the following dimensions: \\[0.5em]

    \begin{itemize}
        \item \textbf{Desired\_Writing\_Style}: Specify the intended tone or narrative voice, such as humorous, formal, poetic, or conversational.
        \item \textbf{Semantic\_Elements}: Clarify the core meaning, focus, or conceptual emphasis that the response should reflect.
        \item \textbf{Morphological\_Constraints}: Indicate any forbidden words, expressions, or formatting (e.g., avoid passive voice or markdown).
        \item \textbf{Multi-lingual\_Constraints}: Specify the language(s) or code-switching rules to be used in the response.
        \item \textbf{Hierarchical\_Instructions}: Define a priority order among multiple tasks, outlining how they should be structured or emphasized.
        \item \textbf{Special\_Output\_Format}: Specify required formats, such as Python code, JSON structure, tables, LaTeX, or HTML.
        \item \textbf{Paragraphs\_Constraints}: Indicate how many paragraphs are needed, and whether any separators (e.g., horizontal lines, ``***'') should be used.
        \item \textbf{Specific\_Sentence}: Require inclusion of a specific sentence at the beginning or end of the response.
        \item \textbf{Key\_Formatting}: Specify formatting of key phrases—such as using \textbf{bold}, \textit{italics}, or ALL CAPS—based on content in the \#Given Output\#.
        \item \textbf{Item\_Listing\_Details}: Define how items should be listed, including use of symbols like bullets (•), numbers (1., 2., 3.), or dashes (-).
    \end{itemize}

    \textbf{\#Given Instruction\#} \\
    \texttt{\{Instruction\}} \\[0.5em]

    \textbf{\#Given Output\#} \\
    \texttt{\{Response\}} \\[0.5em]

    Please format your response directly in JSON, using \texttt{"Constraint\_Type"} as the key and the specific constraint as its value. Ensure that each constraint is a concise and complete sentence of 10–20 words, and use varied phrasing across types.\\[0.5em]

    If a specific type of constraint cannot be derived from the \#Given Output\#, assign the value \texttt{"NULL"}. For example: \texttt{"Constraint\_Type": "NULL"},\\[0.5em]

    Do not include any headings or prefixes in your response. \\
    \bottomrule
    \end{tabular}
    \end{adjustbox}
    \caption{Prompt for generating format, content, and style constraint types based on the back-translation method.}
    \label{tab:constraint_generation_prompt}
\end{table*}

\begin{table*}[t]
    \centering
    \small
    \begin{adjustbox}{max width=\linewidth}
    \begin{tabular}{p{\linewidth}}
    \toprule
    \textbf{Prompt: Extracting Constraints from Instruction} \\[0.5em]
            You are an expert in natural language processing and constraint checking. Your task is to analyze a given instruction and identify which constraints need to be checked.

        The `instruction' contains a specific task query along with several explicitly stated constraints. Based on the instructions, you need to return a list of checker names that should be applied to the constraints.

        [Task Example 1]  \\
        Instruction: Write a 300+ word summary of the Wikipedia page "https://en.wikipedia.org/wiki/Raymond\_III\_Count\_of\_Tripoli". Do not use any commas and highlight at least 3 sections that have titles in markdown format, for example *highlighted section part 1*, *highlighted section part 2*, *highlighted section part 3*.  \\
        
        Response: NumberOfWordsChecker: 300+ word <sep> HighlightSectionChecker: highlight at least 3 sections that have titles in markdown format <sep> ForbiddenWordsChecker: Do not use any commas. \\

        \textbf{\#Task Instruction\#} \\
        \texttt{\{Instruction\}} \\[0.5em]

        \#\#\# Your task: \\
        - Generate the appropriate checker names with corresponding descriptions from the original instruction description.
        - Return the checker names with their descriptions separated by <sep>. \\
        - Focus only on the constraints explicitly mentioned in the instruction. \\
        - Ensure that each constraint is complete, such as specifying whether the 300-word limit applies to the entire text or a specific section. A defined scope is required.\\
        - Do **not** generate checkers for the task query itself or its quality.\\
        - If the instruction is in Chinese/English, please output the constraint in the same language.\\
        - Each checker should be responsible for checking only one constraint.\\
        - Do not output any constraints that are not included in the instruction.\\
        \midrule
         \textbf{Prompt: Classifying Constraints} \\[0.5em]
         Please classify whether the given checker can be judged using simple lexical rules.

         \textbf{\#Checker\#} \\
        \texttt{\{checker\_name\}} \\[0.5em]

        Classification rules: \\
        - If the checker can be determined using simple lexical rules—such as word count, text length, number of paragraphs, number of sentences, or presence of specific keywords—output [[A]]. \\
        - If the checker requires semantic understanding—such as style, tone, sentiment, language, context, genre, or structure—and thus necessitates an additional semantic analysis model (e.g., a large language model), output [[B]]. \\
        - If the constraint is meaningless, non-informative, or irrelevant (e.g., "NA"), output [[C]]. \\
        \midrule
        \textbf{Prompt: Generating code} \\[0.5em]

                You are tasked with implementing a `Python' function `check\_following' that determines whether a given `response' satisfies the constraint defined by a checker. The function should return `True' if the constraint is satisfied, and `False' otherwise. \\
        
        [Example Input 1] \\
        no more than 800 words \newline
        [Example Output 1] \\
        def check\_following(response):    return len(response.split()) <= 800 \\

        [Example Input 2] \\
        Include keywords 'cloud storage', 'open-source' \newline
        [Example Output 2] \\
        import re def check\_following(response):    return bool(re.search(r'cloud storage', response, re.IGNORECASE) and re.search(r'open-source', response, re.IGNORECASE)) \\

        [Example Input 3] \\ 
        The word 'huge' should appear 3 times \newline
        [Example Output 3] \\
        import re def check\_following(response):    return len(re.findall(r'huge', response, re.IGNORECASE)) == 3 \\
        
     \textbf{\#Task Input Checker\#} \\
        \texttt{\{checker\_name\}} \\[0.5em]

        [Requirements] \\
        - The function should be self-contained with necessary imports.  \\
        - DO NOT use nltk. \\
        - Only return exactly `Python` code script, without any other info. \\
    \bottomrule
    \end{tabular}
    \end{adjustbox}
    \caption{Prompt for extracting constraints from instruction, classifying constraint types, and generating code for hard constraints.}
    \label{tab:classify_constraint}
\end{table*}

\end{document}